\documentclass[runningheads]{llncs}
\usepackage[T1]{fontenc}
\usepackage{graphicx}
\usepackage{booktabs}
\usepackage{multirow}
\usepackage{amsmath}
\usepackage{amssymb}
\usepackage{pifont}
\usepackage{adjustbox}
\usepackage{subcaption}

\begin{document}
\title{ProtoSAM: One Shot Medical Image Segmentation With Foundational Models}
\titlerunning{ProtoSAM}
%
\author{Lev Ayzenberg\inst{1},
Raja Giryes\inst{1},
Hayit Greenspan\inst{1,2}}
\authorrunning{L. Ayzenberg et al.}
%
\institute{Faculty of Engineering, Tel Aviv University\and
Icahn school of Medicine, Mount Sinai, NY}
\maketitle              
\begin{abstract}
This work introduces a new framework, ProtoSAM, for one-shot medical image segmentation. It combines the use of prototypical networks, known for few-shot segmentation, with SAM – a natural image foundation model. The method proposed creates an initial coarse segmentation mask using the ALPnet prototypical network, augmented with a DINOv2 encoder. Following the extraction of an initial mask, prompts are extracted, such as points and bounding boxes, which are then input into the Segment Anything Model (SAM). State-of-the-art results are shown on several medical image datasets and demonstrate automated segmentation capabilities using a single image example (one shot) with no need for fine-tuning of the foundation model. Our code is available at: \url{https://github.com/levayz/ProtoSAM/}

\end{abstract}
\section{Introduction}
Deep learning models are now the go-to method for segmenting medical images. However, using these models usually requires a large set of manually labeled data, which is expensive and time-consuming. Additionally, these models struggle with new categories they have not seen before, requiring more training and adjustments.
Few-shot segmentation (FSS) has been introduced as a solution to these problems. FSS trains the model to learn from just a few labeled examples, reducing the need for large, manually labeled datasets.
Prototypical Networks (PN) \cite{snell2017prototypical} are widely used for FSS. These networks use prototypes, which represent the key features of different classes, to make predictions based on similarity. One notable method is ALPNet~\cite{alpnet,alpnet_bp}, which is the leading FSS approach for medical data. Its key innovations are the Adaptive Local Prototypes Pooling module (ALP), which helps to better capture fine details in medical images, and utilizing superpixels as labels for training.

Recently, a segmentation foundational model, Segment Anything Model (SAM) \cite{kirillov2023sam}, was introduced. This model was trained on a curated dataset of 1B masks and 11M natural images. In our work, we aim to implement SAM for medical image segmentation. We hypothesize that with appropriately designed prompts, SAM could serve as an effective one-shot segmentation model for medical images, as we show in Section~\ref{sec:res}. We demonstrate the performance of ProtoSAM on the CHAOS \cite{CHAOS} and MICCAI 2015 Multi-Atlas Abdomen Labeling challenge \cite{miccai2015} datasets and show SOTA results. We also demonstrate the advantages of our strategy on several polyp datasets.

\begin{figure}[t]
\includegraphics[trim=75 0 0 0, clip, width=\textwidth]{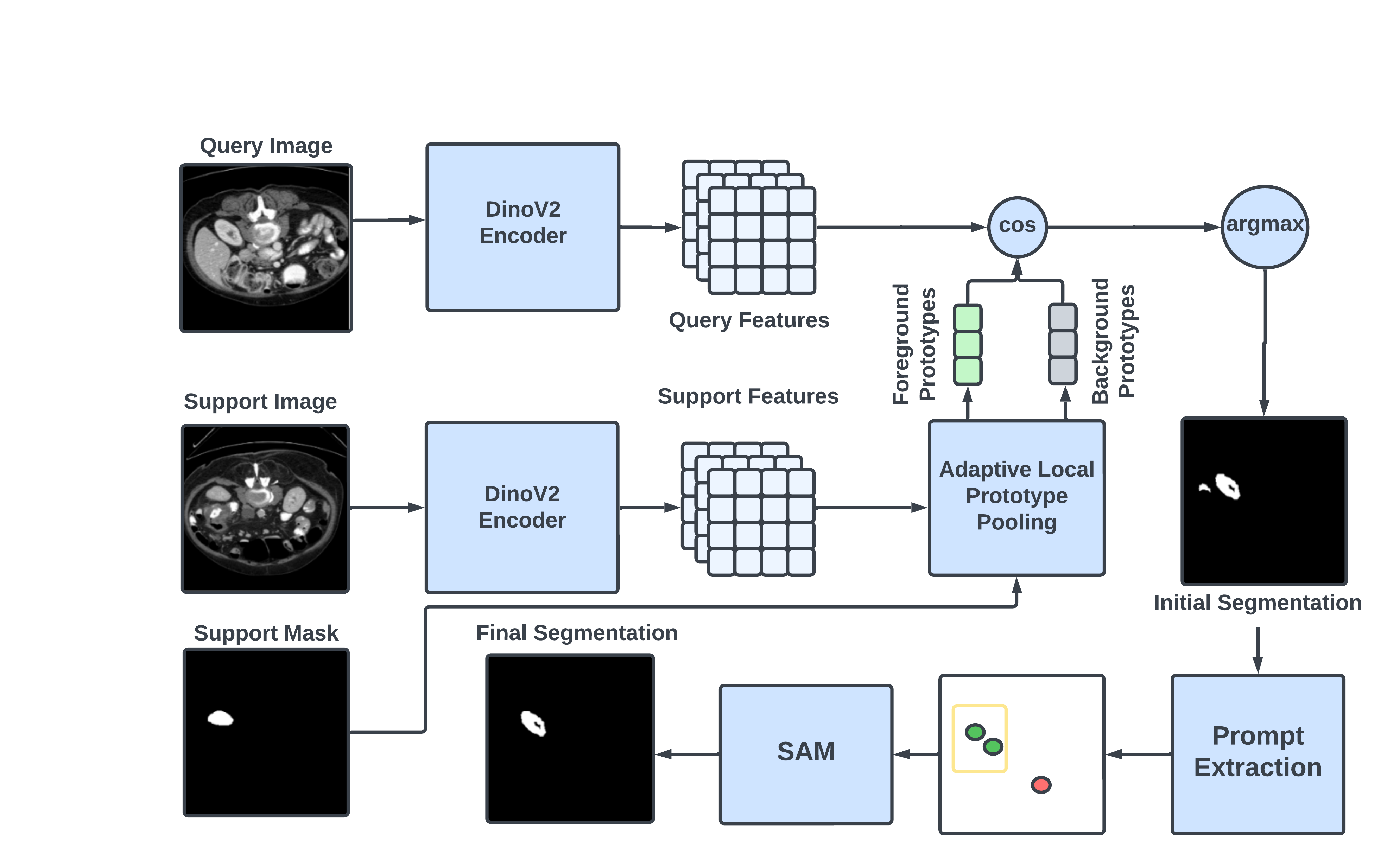}
\caption{ProtoSAM Framework: A DINOv2 encoder derives features from query and support images. Foreground and background prototypes are crafted from support features and masked through the ALP module. Initial segmentation is achieved by comparing these prototypes with query features using cosine similarity. The system extracts prompts from an initial prediction to guide the SAM model for enhanced segmentation.}
\label{fig:prtosam}
\end{figure}

\section{Related Work}
\noindent {\bf Few Shot Segmentation.}
In the supervised setting of medical image segmentation, deep learning models are trained to predict labels for each pixel in the input images. This process typically starts from scratch for each new task, requiring extensive annotated datasets. Few-shot segmentation offers a more practical solution in this context, enabling models to perform well with limited annotated data, typically referred to as the support set \cite{feyjie2020semisupervised,Kotia2021}. 
Prototypical Networks (PN) \cite{snell2017prototypical} are a popular choice for addressing few-shot learning tasks. They focus on exploiting representation prototypes of semantic classes extracted from the support. These prototypes are utilized to make similarity-based predictions. \\
\noindent \textbf{ALPNet} \cite{alpnet,alpnet_bp} is a novel PN approach that introduced the Adaptive Local Prototypes (ALP) module, which is capable of generating prototypes at both local and class-wide levels. Additionally, they proposed a self-supervised learning strategy that employs superpixels as labels for training in a supervised manner.\\
\noindent \textbf{Foundational Models} are large-scale, pre-trained models that serve as a base for various downstream tasks, typically offering strong generalization capabilities. Recently, several vision foundational models have been introduced, demonstrating significant advancements in the field \cite{Rani2023SSL}. \textbf{DINOv2}~\cite{dinov2} is a recent Self Supervised Learning framework based on vision transformers (ViT)~\cite{vit}. It was trained on a large curated dataset of natural images and exhibits strong feature extraction capabilities. Throughout this paper, the term "DINOv2" refers to the ViT that has been pretrained using the DINOv2 framework. In \cite{Ayzenberg2024DINOv2}, the advantage of substituting the encoder in the ALPNet framework with a DINOv2 encoder was demonstrated. In this work, we take an alternative strategy and use ALPNet with DINOv2 for box and point proposals for the Segment Anything Model (\textbf{SAM})  \cite{kirillov2023sam}, which is a foundational model for segmentation. SAM was trained on a large curated set consisting of 1B masks and 11M natural images. SAM supports fully automatic and prompt-based (points and bounding boxes) segmentation. Various adaptation schemes of SAM for medical data have been proposed \cite{shaharabany2023autosam,Zhang2024SAM,wu2023medical}. Recently, MedSAM was introduced \cite{medsam}, which fine-tuned the SAM model for medical images on over 1 million image-mask pairs across 15 different modalities. Yet, unlike our work, MedSAM requires getting bounding box prompts.

In \cite{Ayzenberg2024DINOv2_isbi}, we have employed DINOv2 with ALPNet, showing that using the features generated by DINOv2 led to better segmentation performance on CT and MRI datasets.  In this work, we extend our earlier work to support also the use of SAM in an automatic manner, requiring only a single support image and mask. Furthermore, we expand our evaluation to include polyp datasets.

\section{Method}
\label{method}
\noindent \textbf{Problem Formulation.}
In one-shot segmentation, the goal is to train a function, $f(\mathbf{x}^q, S)$, capable of generating a binary mask for an unseen class $c$ when provided with a support set, $S$, and a query image, $\mathbf{x}^q$.
The support set $S$, comprises pairs of images and their annotated masks ${(\mathbf{x}^s_i(c), \mathbf{y}^s_i(c))_{i=1}^k, c\in C_{test}}$. Since we are dealing with one-shot learning, we set $k=1$. Here, $\mathbf{x}^s$ represents the image in the support image set, and $\mathbf{y}^s(c)$ represents the annotated mask of the support image corresponding to the unseen class $c$.
For training, we split the data into a training dataset, $D_{train}$, and a testing dataset, $D_{test}$. These datasets are comprised of image-binary mask pairs, with $D_{train}$ annotations belonging to $C_{train}$ and $D_{test}$ annotations belonging to $C_{test}$.
It is important to note that $C_{train} \cap C_{test} = \emptyset$.

The training process of FSS networks is comprised of episodes. An episode consists of a pair of a support set and a query image, $\langle S, \mathbf{x}^q \rangle$, where $S$ is a subset of the training dataset $D_{train}$ ($S \subset D_{train}$), and $\mathbf{x}^q$ is a query image. It is crucial to note that the query mask $\mathbf{y}^q(c)$ is utilized solely for training purposes as a ground-truth label. Each episode is randomly selected from $D_{train}$, with the support set $S$ containing   an image-mask pair for a specific class $c$. Given that the problem involves $n$ classes from $C_{test}$, each episode is an "n-way k-shot segmentation sub-problem." In this work, we focus on 1-way 1-shot learning. 

Figure~\ref{fig:prtosam} provides an overview of ProtoSAM. We turn to detail each stage:

\noindent \textbf{Initial Segmentation Stage.} For the initial segmentation stage we employ the ALPNet \cite{alpnet} framework, but we replace the encoder with a DINOv2 \cite{dinov2} encoder.
The purpose of this stage is to extract a coarse segmentation mask and logits from the query image, based on the support set. We encode the support and query image using the encoder, $f_\theta(.)$ and extract the support feature map, $f_{\theta}(\mathbf{x}^s) \in \mathbb{R}^{D\times H\times W}$, and the query feature map $f_{\theta}(\mathbf{x}^q) \in \mathbb{R}^{D\times H\times W}$, where $D$ is the number of channels, and $H,W$ are the feature map spatial dimensions.
The ALP module extracts local prototypes from $f_{\theta}(\mathbf{x}^s)$ by average pooling features for the support feature map using a sliding window. We define a local prototype at location $(m,n)$  for class $c$ as:
\begin{equation}
\label{eq:proto_calc}
    P(m,n;c) = \frac{1}{L_H L_W} \sum_{h} \sum_{w} f_{\theta}(\mathbf{x}^s(c)(h,w),
\end{equation}
where $(L_H, L_W)$ are the sliding window dimensions and  $mL_H \leq h < (m+1) L_H, \ nL_W \leq w < (n+1)L_W$.

Additionally, we extract a global prototype defined as:
\begin{equation}
    P^g(c) = \frac{\sum\limits_{h,w}\mathbf{y}_i^s(c)(h,w)f_{\theta}(\mathbf{x}^s(c))(h,w)}{\sum\limits_{h,w}\mathbf{y}^s(c)(h,w)}
    \label{eq:class_level_proto}
\end{equation}
For cases where the foreground class may be much smaller than the pooling window.
For each class (foreground or background), the prototypes are aggregated into a set $\mathbb{P}_c = \{P_l(c)\}$.
A  similarity map is generated between each prototype of class $c$  and the query feature map, $f_{\theta}(\mathbf{x}^q)$, using:
\begin{equation}
    S_{l}(c^j)(h, w) = \alpha\cdot P_{l}(c)\odot f_{\theta}(\mathbf{x}^q)(h,w)
    \label{eq:local_sim_map}
\end{equation}
Where $\odot$ denotes cosine similarity, and $l$, the index for the prototype. We set $\alpha=20$ as in \cite{alpnet}.
These similarity maps are merged into a class-wise similarity map using:
\begin{equation}
S'(c)(h,w) = \underset{l}{\sum} S_l(c)(h,w) \cdot \underset{l}{\text{softmax}}[S_l(c)(h,w)]
\label{eq:class_sim}
\end{equation}
We normalize the similarity maps for the foreground and background class into probabilities using:
\begin{equation}
\label{equ: pred_prob}
\mathbf{\hat{y}}^q(h,w) = \underset{c}{\text{softmax}}[S'(c)(h,w) ]
\end{equation}
It is important to note that we do not perform any fine-tuning unless explicitly stated otherwise, and rely solely on the feature extraction capabilities of the DINOv2 encoder.

\noindent \textbf{Prompt Extraction}. This stage receives as input the initial prediction probabilities. From these, we extract prompts in the form of points and bounding boxes for SAM \cite{kirillov2023sam}. 
Using connected component analysis, we first extract the most confident connected component from the initial prediction using the following formula to rank the prediction confidence of each connected component:
\begin{equation}
\text{Confidence} = \frac{\sum_{i} p_{i} \cdot \mathbf{\hat{y}}_i}{\sum_{i} \mathbf{\hat{y}}_i}
\label{eq:cca_conf}
\end{equation}
where $p_i$ is the probability that pixel $i$ belongs to the foreground class, $\mathbf{\hat{y}}_i$ is the $i$-th pixel's predicted label. From this point on, we only consider the probabilities and mask corresponding to that connected component.
We extract the following prompts from the predicted logits and mask: bounding box (Bbox), centroid points (Cent) and confidence-based points (Conf).
We set the bounding box to be the bounding box of the connected component, the centroid point is selected to be the centroid point of the connected component, and for the confidence-based point, we choose the point corresponding to the highest probability of the foreground class inside the connected component.

\noindent \textbf{Final Segmentation}. The prompts of the previous stage are fed into the Segment Anything Model (SAM) which produces the final finer segmentation map.

\noindent \textbf{Optional: Encoder fine-tuning (EFT).} To enhance performance, it is possible to fine-tune the DINOv2 encoder on the target data. For that, we employ the ALPNet framework \cite{alpnet}. We emulate real-life scenarios by constructing episodes, comprised of a support set and a query set. A preprocessing step is taken, where we generate superpixels using Felzenszwalb \cite{felzenszwalb2004efficient} for all the available slices.
At each episode an image $\mathbf{x}$ is chosen from the train-set, together with a random superpixel $\mathbf{y}^r(c^p)$, which is used as the pseudo-label. This forms the support set $S={(\mathbf{x}, \mathbf{y}^r(c^p))}$, where $c^p$ denotes the superpixel class and $r$ is the index of the random superpixel. The query set is formed by augmenting the chosen image $\mathbf{x}$, i.e, $Q_i={(\mathcal{T}_g(\mathcal{T}_I(\mathbf{x}}))$, where $\mathcal{T}_g$ and $\mathcal{T}_I$ are geometric and intensity transforms respectively. The loss is
\begin{eqnarray}
\label{equ:loss_seg}
 \mathcal{L}^i_{\text{seg}}(\theta ; S, Q) = 
- \frac{1}{HW} \sum_{h=1}^H \sum_{w=1}^W \sum_{c \in \{c^0, c^p\}} \mathcal{T}_g(\mathbf{y}^r(c))(h,w) \log( \hat{\mathbf{y}}^r(c)(h,w) ),
\end{eqnarray}
where $\hat{\mathbf{y}}^r(c^p)$ is the prediction of the pseudolabel, $c^0$ is the background class and $\theta$ - the model parameters. Also, as in \cite{alpnet}, we incorporate the \textit{prototype alignment regularization} \cite{wang2020panet}, where the roles of the support label and the prediction are reversed. The prediction assumes the role of the support label, and our aim is to segment the original superpixel accordingly.
The regularization loss is
\begin{eqnarray}
\label{equ:loss_reg}
\mathcal{L}_{\text{reg}}(\theta ; \mathcal{S}', \mathcal{S}) =
- \frac{1}{HW} \sum_{h=1}^H \sum_{w=1}^W \sum_{c \in \{c^0,c^p\} } \mathbf{y}^r(c)(h,w) \log( \bar{\mathbf{y}}^r(c)(h,w) ),
\end{eqnarray}
where $S'=(\mathbf{x}, \hat{\mathbf{y}}^r(c^p))$, and  $\bar{\mathbf{y}}^r(c^p)$ is the prediction of the superpixel label $\mathbf{y}^r(c^p)$.

\section{Experiments}
\noindent \textbf{Datasets.} We employ two datasets for abdominal organ segmentation, each associated with a different modality (CT and MRI), and four polyp datasets. The first two datasets include (i) Abd-CT: Derived from the MICCAI 2015 Multi-Atlas Abdomen Labeling challenge \cite{miccai2015}, containing 30 3D abdominal CT scans and (ii) Abd-MRI: Sourced from the ISBI 2019 Combined Healthy Abdominal Organ Segmentation Challenge (Task 5) \cite{CHAOS}, comprising 20 3D T2-SPIR MRI scans.
The poylp datasets are: Kvasir-SEG \cite{jha2019kvasirseg}, ClinicDB \cite{cvcclinicdb}, ColonDB \cite{cvccolondb} and ETIS \cite{etis}. The test splits of each dataset are as follows: 100 images from ClinicDB, 64 from Kvasir-SEG, 380 from ColonDB, 196 from ETIS, following \cite{shaharabany2023autosam}. 

\noindent \textbf{Evaluation.}
To evaluate 2D segmentation on 3D volumetric images, we follow the evaluation protocol established by \cite{squeeze_and_excite}.
In a 3D image, when dealing with each specific class denoted as $c^{j}$, we divide the images that fall between the top and bottom slices containing this class into equal sections. These sections, in our experiments, are set to be $C=3$ in number. For each of these sections, we choose the middle slice from the corresponding section of the support scan as a reference point. Then, this reference slice is used to guide the segmentation of all the slices within the current section in the query scan. It is important to note that the support and query scans are obtained from different scans.
We employ Dice score and Intersection over Union (IoU) as evaluation metrics. The Dice score, used for the CT and MRI datasets, is defined as follows:
\begin{equation}
\text{Dice} = \frac{2 \cdot |A \cap B|}{|A| + |B|}
\end{equation}
where \(A\) and \(B\) are the sets of predicted and ground truth binary segmentation masks, respectively.

The IoU score, used alongside the Dice score for the polyp datasets, is defined as follows:
\begin{equation}
\text{IoU} = \frac{|A \cap B|}{|A \cup B|}
\end{equation}
For the CT and MRI datasets, we use the Dice score since IoU was not available for comparison with other models. For the polyp datasets, we use both the Dice score and IoU.
For the CT and MRI evaluation, we perform 5-fold cross-validation for each task. The results, including the mean and standard deviation, are presented in Table \ref{table:merged_ct_mri_v2}.
Additionally, we perform a statistical analysis using the Wilcoxon rank test (Table \ref{table:wilcoxon}) using compiled results for all organs per modality (4 organs × 5 folds = 20 data points per model). Some models were excluded due to missing fold results.

\noindent \textbf{Implementation Details.}
Our code\footnote{\url{https://github.com/levayz/ProtoSAM/}} is freely available.
For inference images are resized to $672 \times 672$ resolution.
For (optionally) fine-tuning the encoder, we employ the ALPNet \cite{alpnet} framework as described in Section \ref{method}, together with Low Rank Adaption (LoRA) \cite{hu2021lora}.  We train for 100k steps with a learning rate of 1e-4 and resize the images to $256 \times 256$. For the CT and MRI datasets, we use a setting introduced in \cite{alpnet}, where the testing class may not appear during training, meaning we discard any slice that contains the testing class. This setting is referred to as ``Setting 2'' in \cite{alpnet}. 

For the MRI and CT datasets, we divide the organs into two groups: (Spleen, Liver), (RK, LK), where RK and LK represent Right-Kidney and Left-Kidney respectively. 
For the DINOv2 encoder we use the 'large' architecture. We use the SAM 'huge' architecture for ProtoSAM, AutoSAM, and SAM models, and the 'base' architecture for ProtoSAM-base and ProtoMedSAM.

\section{Results}
\label{sec:res}

\noindent \textbf{Comparison to SOTA methods.} Our method was benchmarked against leading methods across different datasets. For the polyp dataset, we compared our model to the SOTA supervised segmentation model, AutoSAM \cite{shaharabany2023autosam}. For fair comparison, we trained AutoSAM with support sets of 1, 5, and 100 images and their masks, before evaluating it.
We also assessed SAM, using its automatic segmentation mode, and manually selecting its best-performing mask. Additionally, we compared ProtoSAM with both the default DINOv2 encoder and a fine-tuned version as described in Section~\ref{method}. Finally, we evaluated the performance of ProtoMedSAM, where we replace SAM with MedSAM in ProtoSAM and use only bounding boxes as prompts, as MedSAM was fine-tuned this way.

\begin{table}[t]
\caption{MRI/CT 1-Shot Results (in Dice score) on abdominal images}
\centering
\resizebox{\linewidth}{!}{
\begin{tabular}{lc|c|c|c|c}
\toprule
\multirow{2}{*}{Method} & \multicolumn{1}{c}{LK} & \multicolumn{1}{c}{RK} & \multicolumn{1}{c}{Spleen} & \multicolumn{1}{c}{Liver} & {Mean} \\
\cline{2-6}
& \multicolumn{1}{c}{MRI / CT} & \multicolumn{1}{c}{MRI / CT} & \multicolumn{1}{c}{MRI / CT} & \multicolumn{1}{c}{MRI / CT} & \multicolumn{1}{c}{MRI / CT} \\
\midrule
SSL-ALPNet \cite{alpnet} & 73.63$\pm$4.60 / 63.34$\pm$9.33 & 78.39$\pm$5.51 / 54.82$\pm$11.7 & 67.02$\pm$7.98 / 60.25$\pm$6.67 & 73.05$\pm$3.49 / 73.65$\pm$3.61 & 73.02$\pm$6.01 / 63.02$\pm$7.97 \\
SSL-ALPNet+BP \cite{alpnet_bp} & 78.77 / 66.04 & 83.44 / 62.14 & 70.02 / 68.39 & 75.01 / 73.90 & 76.81 / 67.62 \\
CRAP-Net \cite{crapnet} & 74.66 / 70.91 & 82.77 / 67.33 & 73.82 / 70.17 & 70.82 / 70.45 & 73.82 / 69.72  \\
CRTPNet \cite{cross_reference_transformer} & 76.74 / 66.37 & 80.15 / 61.05 & 70.07 / 67.92 & 73.36 / 73.88 & 75.08 / 67.31 \\
AutoSAM \cite{shaharabany2023autosam} & 61.07 / 43.20 & 64.46 / 38.77 & 69.03 / 54.50 & 68.10 / 70.68 & 65.66 / 51.79 \\
SAM (best mask) \cite{kirillov2023sam} & 77.32$\pm$4.54 / \textbf{85.21}$\pm$3.29 & 80.75$\pm$3.85 / \textbf{85.36}$\pm$2.63 & 66.37$\pm$4.25 / 76.56$\pm$4.44 & 27.61$\pm$6.61 / 69.58$\pm$1.39 & 63.01$\pm$1.39 / \textbf{79.18}$\pm$3.29 \\
SSL-DINOv2 \cite{Ayzenberg2024DINOv2} &  75.06$\pm$11.61 / 69.96$\pm$11.61 & 80.21$\pm$2.14 / 66.40$\pm$12.99 & 71.86$\pm$10.30 / 73.00$\pm$6.49 & 73.50$\pm$5.95 / 76.40$\pm$3.89 & 75.16$\pm$2.14 / 71.44$\pm$6.49 \\
SSL-DINOv2+CCA\cite{Ayzenberg2024DINOv2} &  81.43$\pm$13.84 / 66.40$\pm$13.84 & 84.40$\pm$4.07 / 69.96$\pm$13.65 & 73.30$\pm$10.27 / 74.60$\pm$6.58 & 74.20$\pm$5.46 / 81.67$\pm$4.41 & 78.43$\pm$4.07 / 73.16$\pm$6.58 \\
\hline
ProtoMedSAM & 69.97$\pm$4.97 / 66.08$\pm$9.45 & 77.16$\pm$3.92 / 67.15$\pm$6.01 & 69.68$\pm$8.38 / 60.53$\pm$2.94 & 71.99$\pm$6.12 / 78.64$\pm$3.03 & 72.20$\pm$4.44 / 68.10$\pm$4.44 \\
ProtoSAM-base &  70.47$\pm$3.93 / 67.54$\pm$8.33 & 79.03$\pm$3.06 / 64.52$\pm$7.85 & 69.56$\pm$7.87 / 57.99$\pm$3.40 & 69.88$\pm$4.89 / 77.56$\pm$2.50 & 72.23$\pm$3.93 / 66.90$\pm$8.33 \\
ProtoSAM & 71.46$\pm$6.85 / 69.44$\pm$10.21 & 81.43$\pm$5.16 / \textbf{71.04$\pm$10.52 } & 76.51$\pm$7.97 / 65.50$\pm$7.72 & \textbf{81.94$\pm$6.83 } / \textbf{87.84$\pm$4.71 } & 77.83$\pm$6.85 / 73.45$\pm$10.21 \\
ProtoSAM +EFT & \textbf{85.61$\pm$3.90} / \textbf{73.88$\pm$16.50} & \textbf{87.94$\pm$1.79} / 70.05$\pm$14.27 & \textbf{80.92$\pm$7.12} / \textbf{81.14$\pm$3.83 } & 80.60$\pm$3.33 / 86.34$\pm$ 3.20& \textbf{83.69}$\pm$3.33 / \textbf{77.85}$\pm$14.27 \\
\bottomrule
\end{tabular}
}
\label{table:merged_ct_mri_v2}
\end{table}

\begin{table}[h]
\centering
\begin{tabular}{lcc}
\toprule
Comparison & CT (p-value) & MRI (p-value) \\
\midrule
ProtoSAM vs ALPNet & 0.0027 & 0.001 \\
ProtoSAM+EFT vs SSL-DINOv2+CCA & 0.0017 & 3.62E-05 \\
\bottomrule
\end{tabular}
\caption{Wilcoxon tests results}
\label{table:wilcoxon}
\end{table}

\begin{figure}[t]
    \centering
        \includegraphics[width=0.45\textwidth]{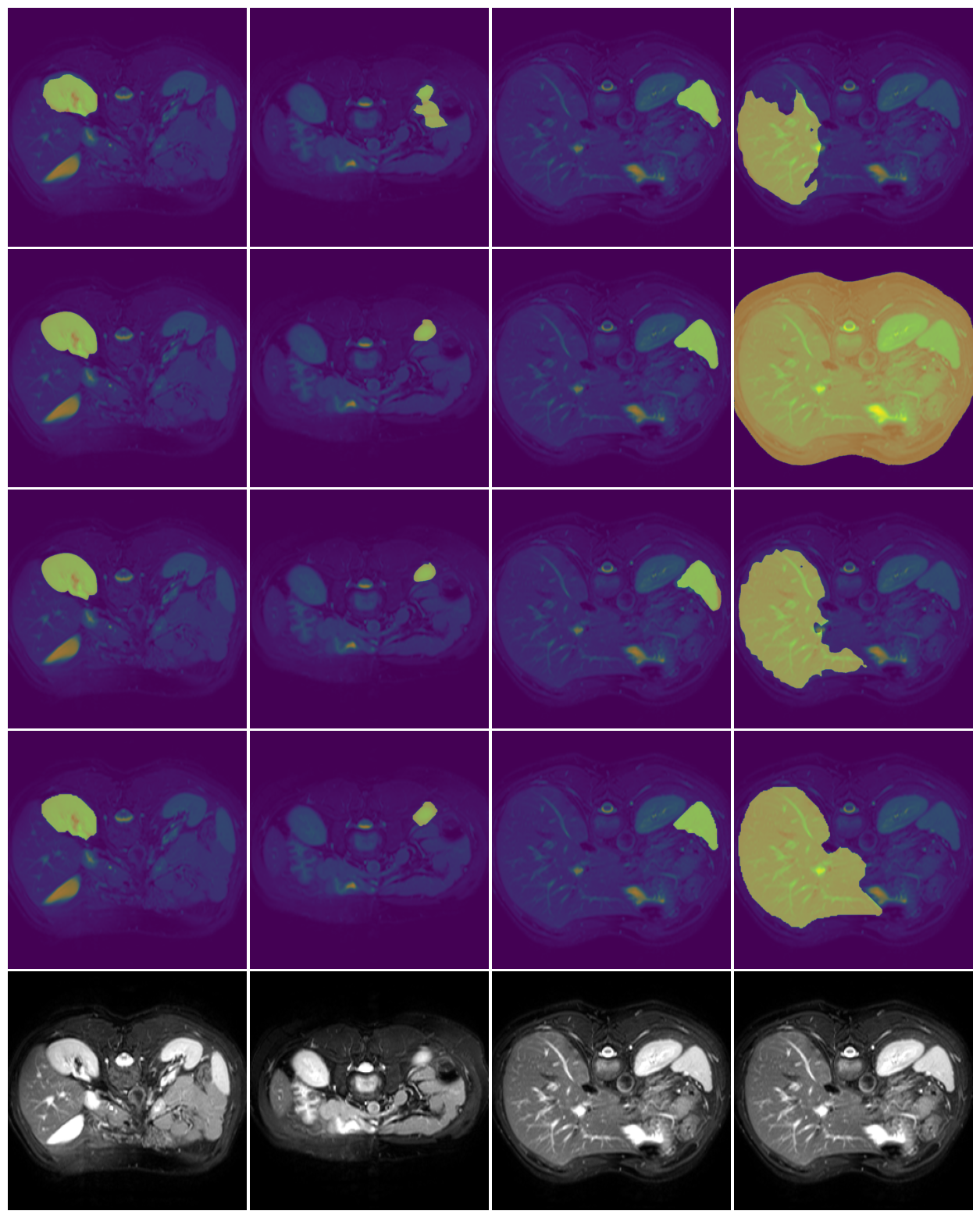}
    \hfill
        \includegraphics[width=0.45\textwidth]{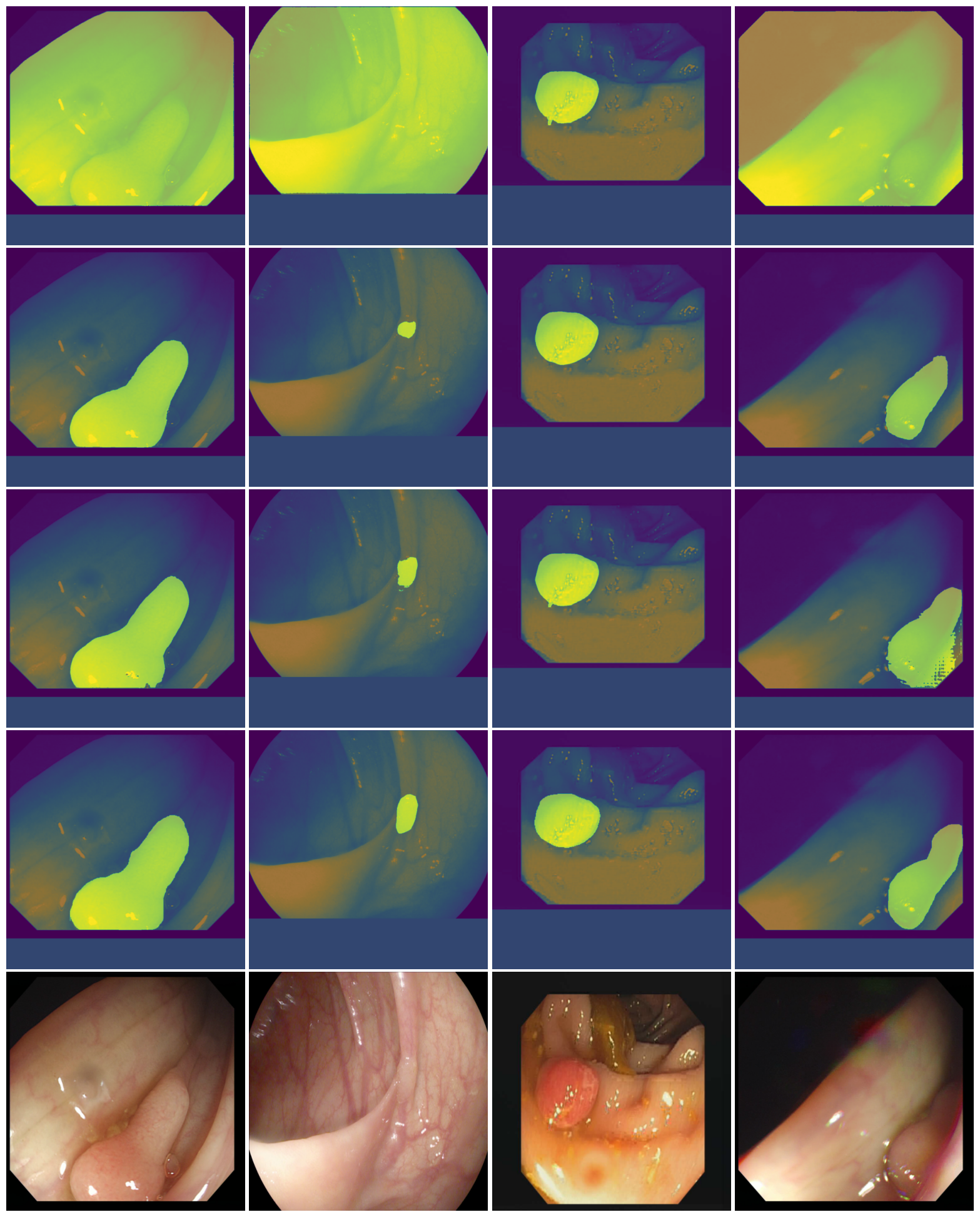}
    \caption{(Left) \textbf{MRI segmentation.} Top to bottom: SSL-Dinov2 + CCA, SAM (best mask), ProtoSAM, Ground Truth, Query Image. Left to right: RK, LK, Spleen, Liver. (Right) \textbf{Polyp segmentation.} Top to bottom: SAM (best mask), ProtoMedSAM, ProtoSAM, Ground Truth, Query Image.}
    \label{fig:both_figures}
\end{figure}

\begin{table}[t]
\caption{Polyp segmentation benchmarks results}
\begin{center}
\resizebox{\linewidth}{!}{
    \begin{tabular}{@{}l@{~}cccccccccc@{}}
    \toprule
    \multirow{2}{*}{Method}&\multicolumn{2}{c}{Kvasir33~\cite{jha2019kvasirseg}}&\multicolumn{2}{c}{Clinic~\cite{cvcclinicdb}}&\multicolumn{2}{c}{Colon~\cite{cvccolondb}}&\multicolumn{2}{c}{ETIS~\cite{etis}}&\multicolumn{2}{c}{All}\\ \cmidrule(l){2-11} 
     & Dice & IoU &  Dice & IoU  & Dice & IoU &  Dice & IoU &  Dice & IoU\\
    \midrule
    AutoSAM~\cite{shaharabany2023autosam} 1-shot & {35.64} & {25.18} & {20.31} & {14.12} & {10.34} & {6.29} & {13.97} & {9.00} & {15.57} & {10.23}\\
    AutoSAM~\cite{shaharabany2023autosam} 5-shot & {59.56} & {48.14} & {38.90} & {29.34} & {35.60} & {25.75} & {26.57} & {19.53} & {36.72} & {33.33}\\
    AutoSAM~\cite{shaharabany2023autosam} 100-shot & {73.44} & {62.51} & {62.64} & {52.26} & {53.27} & {43.24} & {38.52} & {29.94} & {52.87} & {43.08} \\
    AutoSAM~\cite{shaharabany2023autosam} full-set & 91.00 & 87.00 & 92.80 & 89.33 & 83.00 & 76.70 & 79.70 & 74.00 & 86.25 & 81.75 \\ 
    SAM (best mask)~\cite{kirillov2023sam} & {77.91} & {69.28} & {57.23} & {50.66} & {46.08} & {39.73} & {52.32} & {47.42} & {59.94} & {53.98} \\
    \midrule
    ProtoMedSAM (1-shot) & 81.11 & 72.75 & 66.54 & 57.31 & 64.34 & 55.61 & 59.899 & 51.66 & 65.62 & 57.02 \\
    ProtoSAM-base (1-shot) & 79.98 & 71.12 & 65.79 & 56.12 & 65.48 & 57.11 & \textbf{63.07} & 55.50 & 66.83 & 58.50 \\
    ProtoSAM (1-shot) & \textbf{81.55} & \textbf{73.66} & \textbf{69.01} & \textbf{60.80} & \textbf{67.31} & \textbf{59.60} & 62.33 & \textbf{55.47} & \textbf{68.06} & \textbf{60.49} \\
    \bottomrule
    \end{tabular}
}
\end{center}
\label{tab:polyp}
\end{table}

\noindent \textbf{Discussion.}
As shown in Table \ref{table:merged_ct_mri_v2}, for CT images, ProtoSAM achieves a mean Dice score of 73.45. It surpasses other methods in the liver region with a score of 87.84, outperforming other trained methods and the best mask produced by the SAM model. However, in the spleen region, its score of 65.50 is lower than the scores of some other methods. Additionally, in certain cases, SAM provides better masks than ProtoSAM, indicating that there may be room for improvement in prompt generation. This suggests that while ProtoSAM is promising, further training and refinement could enhance its performance in specific areas. Note though that in SAM, we select the best mask w.r.t GT, which is not done in ProtoSAM.

Table \ref{table:wilcoxon} presents results of the paired Wilcoxon tests. These results indicate significant differences between the models. For ProtoSAM vs ALPNet, the p-values were 0.0027 for CT and 0.001 for MRI. For ProtoSAM EFT vs SSL-DINOv2+CCA, the p-values were 0.0017 for CT and 3.62E-05 for MRI, demonstrating statistically significant improvements in both cases.

For MRI, ProtoSAM outperforms previous trained methods in the Liver, Spleen organs, as well as in the overall mean, with scores of 81.94, 76.51 and 77.83, respectively. It is also better than SAM on average in this case.
By applying the EFT described in Section~\ref{method}, we observe a boost in performance, enabling ProtoSAM to surpass all previous methods on all benchmarks, excluding SAM (that does best mask selection w.r.t GT) on LK and RK in CT.

Table~\ref{tab:polyp} presents the results of the polyp segmentation benchmarks to compare the performance of ProtoSAM (1-shot) with SAM and AutoSAM (under different shot settings, including training on the full training dataset). The evaluation metrics used are the Dice coefficient and Intersection over Union (IoU).
ProtoSAM (1-shot) achieves a mean Dice score of 68.06 and a mean IoU of 60.49 across all datasets. This performance surpasses that of SAM when choosing its best mask w.r.t GT. This indicates that our prompt generation process is effective.

Observe that ProtoSAM consistently outperforms ProtoMedSAM across all our test cases. Note that the ProtoSAM with SAM-base model provided comparable performance to ProtoMedSAM for MRI and CT but better results for the polyps. We believe the gap stems from the fact that MedSAM was fine-tuned using bounding boxes derived from ground-truth masks, ensuring high accuracy in the targeted regions. Since the ProtoSAM framework relies on an initial coarse segmentation, the bounding boxes are not as accurate. In contrast, SAM in ProtoSAM employs both bounding boxes and points. Since it was not fine-tuned with precise bounding boxes, it is more resilient to Bbox noise.

When comparing ProtoSAM (1-shot) to AutoSAM under different shot settings, it is evident that our untrained method achieves better performance using fewer images. Specifically, ProtoSAM (1-shot) outperforms AutoSAM (1-shot), AutoSAM (5-shot), and even AutoSAM (100-shot) in terms of mean Dice and IoU scores, despite using only one image. This highlights the efficiency of ProtoSAM in leveraging pre-training to achieve superior segmentation results compared to the best supervised framework on these datasets, namely AutoSAM. However, there remains a performance gap between the fully-supervised model and the 1-shot performance of ProtoSAM.
Using EFT with Felzenszwalb generated superpixels for poylps, lead to degregation of results compared to using the pretrained encoder. This is a topic for future research.

\noindent \textbf{Ablation Study.}
We wish to examine the effects of using different prompt extraction methods.
We tested the effects of the prompts on all the polyp segmentation datasets and the first evaluation fold from the MRI dataset.
Table~\ref{tab:ablation_prompts} present the results of this study on polyp segmentation datasets and the first evaluation fold of the MRI dataset, respectively. In these tables, we compare various combinations of prompts, including centroid (Cent), confidence (Conf), negative (Neg) point, and bounding box (Bbox).
Negative points are generated in the same way as Conf points, only for the background class instead of the foreground class.
Our findings indicate that the combination of bounding box, confidence, and centroid prompts (Bbox+Conf+Cent) yielded the best overall performance across both datasets.

\begin{table}[t]
\caption{Ablation study of different prompts}
\label{tab:ablation_prompts}
\begin{center}
\resizebox{\linewidth}{!}{
    \begin{tabular}{cccc|cc|cc|cc|cc|cc|cc}
    \toprule
    \multirow{2}{*}{Cent} & \multirow{2}{*}{Conf}  & \multirow{2}{*}{Neg} & \multirow{2}{*}{Bbox} & \multicolumn{2}{|c|}{MRI LK} & \multicolumn{2}{c|}{MRI RK} & \multicolumn{2}{|c|}{MRI Spleen} & \multicolumn{2}{c|}{MRI Liver} & \multicolumn{2}{c|}{Polyps} & \multicolumn{2}{c}{\textbf{Mean}}\\
    & & & & \multicolumn{1}{|c}{Dice} & \multicolumn{1}{c|}{IoU} & \multicolumn{1}{c}{Dice} & \multicolumn{1}{c|}{IoU} & \multicolumn{1}{c}{Dice} & \multicolumn{1}{c|}{IoU} & \multicolumn{1}{c}{Dice } & \multicolumn{1}{c|}{IoU} & \multicolumn{1}{c}{Dice} & \multicolumn{1}{c|}{IoU} & \multicolumn{1}{c}{Dice} & \multicolumn{1}{c}{IoU} \\
        \midrule
    \ding{51} & \ding{55} & \ding{55} & \ding{55} & 56.31 & 51.51 & 77.11 & 72.22 & 72.67 & 67.29 & 47.86 & 36.29 & 55.38 & 48.33 & 61.87 & 55.13 \\
    \ding{55} & \ding{51} & \ding{55} & \ding{55} & \textbf{65.60} & \textbf{60.92} & 78.82 & 73.98 & \textbf{75.55} & \textbf{69.60} & 35.67 & 27.81 & 58.73 & 51.32 & 62.87 & 56.73 \\
    \ding{51} & \ding{51} & \ding{55} & \ding{55} & 60.76 & 55.42 & 80.11 & 75.04  & 73.46 & 68.00 & 53.59 & 43.00 & 62.32 & 54.89 & 66.05 & 59.27 \\
    \ding{55} & \ding{55} & \ding{55} & \ding{51} & 62.42 & 57.38 & 82.40 & 76.58 & 74.61 & 68.43 & 77.37 & 67.56 & 67.43 & 59.80 & 72.85 & 65.95 \\
    \ding{51} & \ding{51} & \ding{55} & \ding{51} & 61.71 & 56.83 & \textbf{82.81} & \textbf{77.27} & 74.00 & 68.08 & \textbf{78.87} & \textbf{69.18} & \textbf{68.06} & \textbf{60.49} & \textbf{73.14} & \textbf{66.42} \\
    \ding{51} & \ding{51} & \ding{51} & \ding{51} & 61.69 & 56.79 & 82.59 & 76.89 & 73.87 & 67.88 & 78.47 & 68.48 & 67.99 & 60.24 & 72.11 & 65.18 \\
    \bottomrule
    \end{tabular}
}
\end{center}
\end{table}

\section{Limitations}
The current method, selects prompts based on the most confident component of the initial segmentation phase, limiting it a single object segmentation, making it unsuitable for tasks requiring multiple object segmentation.
Additionally, while ProtoSAM outperforms many existing methods, there are still cases where SAM provides better masks, indicating room for improvement in the prompt generation process. Furthermore, there is a performance gap between fully-supervised models and ProtoSAM's one-shot performance,  evident in the polyp segmentation results, suggesting that further refinement of the method is needed to match SOTA supervised approaches. Lastly, the encoder fine-tuning (EFT) technique, which improved results for CT and MRI datasets, led to degradation of results for polyp segmentation when using Felzenszwalb generated superpixels, highlighting the need for a more robust fine-tuning strategy across diverse datasets.

\section{Conclusion}
In this work we introduce ProtoSAM,  a novel automatic framework for one-shot image segmentation. The experimental results demonstrate that the untrained ProtoSAM achieves competitive performance across various organs and modalities, often surpassing trained methods. Encoder fine-tuning further boosts its performance, enabling it to outperform all previous methods, including the best masks generated by the original SAM model, in specific areas. Future work could explore the refinement of prompt generation, including prompt generation for MedSAM and extension of the framework to other segmentation tasks. To conclude, ProtoSAM presents a promising approach to one-shot image segmentation in medical imaging, with potential applications in scenarios where labeled data is scarce or where rapid adaptation to new classes is required. 

\begin{credits}
\subsubsection{\ackname} The research in this publication was supported in part by the Israel Science Foundation (ISF) grant number 20/2629, the Israel Ministry of Science and Technology, KLA research fund, and TAD.

\end{credits}
%
%
%
\bibliographystyle{splncs04}
\bibliography{protosam}

\begin{thebibliography}{10}
\providecommand{\url}[1]{\texttt{#1}}
\providecommand{\urlprefix}{URL }
\providecommand{\doi}[1]{https://doi.org/#1}

\bibitem{Ayzenberg2024DINOv2_isbi}
Ayzenberg, L., Giryes, R., Greenspan, H.: Dinov2 based self supervised learning
  for few shot medical image segmentation. In: 2024 IEEE 21st International
  Symposium on Biomedical Imaging (ISBI). IEEE (2024)

\bibitem{Ayzenberg2024DINOv2}
Ayzenberg, L., Giryes, R., Greenspan, H.: Dinov2 based self supervised learning
  for few shot medical image segmentation. arXiv preprint arXiv:2403.03274
  (2024)

\bibitem{cvcclinicdb}
Bernal, J., S{\'a}nchez, F.J., Fern{\'a}ndez-Esparrach, G., Gil, D.,
  Rodr{\'\i}guez, C., Vilari{\~n}o, F.: Wm-dova maps for accurate polyp
  highlighting in colonoscopy: Validation vs. saliency maps from physicians.
  Computerized medical imaging and graphics  \textbf{43},  99--111 (2015)

\bibitem{crapnet}
Ding, H., Sun, C., Tang, H., Cai, D., Yan, Y.: Few-shot medical image
  segmentation with cycle-resemblance attention. In: Proceedings of the
  IEEE/CVF Winter Conference on Applications of Computer Vision. pp. 2488--2497
  (2023)

\bibitem{vit}
Dosovitskiy, A., Beyer, L., Kolesnikov, A., Weissenborn, D., Zhai, X.,
  Unterthiner, T., Dehghani, M., Minderer, M., Heigold, G., Gelly, S., et~al.:
  An image is worth 16x16 words: Transformers for image recognition at scale.
  arXiv preprint arXiv:2010.11929  (2020)

\bibitem{felzenszwalb2004efficient}
Felzenszwalb, P.F., Huttenlocher, D.P.: Efficient graph-based image
  segmentation. International journal of computer vision  \textbf{59},
  167--181 (2004)

\bibitem{feyjie2020semisupervised}
Feyjie, A.R., Azad, R., Pedersoli, M., Kauffman, C., Ayed, I.B., Dolz, J.:
  Semi-supervised few-shot learning for medical image segmentation (2020)

\bibitem{hu2021lora}
Hu, E.J., Shen, Y., Wallis, P., Allen-Zhu, Z., Li, Y., Wang, S., Wang, L.,
  Chen, W.: Lora: Low-rank adaptation of large language models. arXiv preprint
  arXiv:2106.09685  (2021)

\bibitem{cross_reference_transformer}
Huang, Y., Liu, J.: Cross-reference transformer for few-shot medical image
  segmentation. arXiv preprint arXiv:2304.09630  (2023)

\bibitem{jha2019kvasirseg}
Jha, D., Smedsrud, P.H., Riegler, M.A., Halvorsen, P., de~Lange, T., Johansen,
  D., Johansen, H.D.: Kvasir-seg: A segmented polyp dataset (2019)

\bibitem{CHAOS}
Kavur, A.E., Gezer, N.S., Bar{\i}{\c{s}}, M., Aslan, S., Conze, P.H., Groza,
  V., Pham, D.D., Chatterjee, S., Ernst, P., Özkan, S., Baydar, B., Lachinov,
  D., Han, S., Pauli, J., Isensee, F., Perkonigg, M., Sathish, R., Rajan, R.,
  Sheet, D., Dovletov, G., Speck, O., Nürnberger, A., Maier-Hein, K.H., Akar,
  G.B., Ünal, G., Dicle, O., Selver, M.A.: {CHAOS} challenge - combined
  ({CT}-{MR}) healthy abdominal organ segmentation. Medical Image Analysis
  \textbf{69},  101950 (apr 2021)

\bibitem{kirillov2023sam}
Kirillov, A., Mintun, E., Ravi, N., Mao, H., Rolland, C., Gustafson, L., Xiao,
  T., Whitehead, S., Berg, A.C., Lo, W.Y., et~al.: Segment anything. arXiv
  preprint arXiv:2304.02643  (2023)

\bibitem{Kotia2021}
Kotia, J., Kotwal, A., Bharti, R., Mangrulkar, R.: Few Shot Learning for
  Medical Imaging, pp. 107--132. Springer International Publishing (2021)

\bibitem{miccai2015}
Landman, B., Xu, Z., Igelsias, J., Styner, M., Langerak, T., Klein, A.: Miccai
  multi-atlas labeling beyond the cranial vault--workshop and challenge. In:
  Proc. MICCAI Multi-Atlas Labeling Beyond Cranial Vault—Workshop Challenge.
  vol.~5, p.~12 (2015)

\bibitem{medsam}
Ma, J., He, Y., Li, F., Han, L., You, C., Wang, B.: Segment anything in medical
  images. Nature Communications  \textbf{15}(1), ~654 (2024)

\bibitem{dinov2}
Oquab, M., Darcet, T., Moutakanni, T., Vo, H.V., Szafraniec, M., Khalidov, V.,
  Fernandez, P., Haziza, D., Massa, F., El-Nouby, A., Howes, R., Huang, P.Y.,
  Xu, H., Sharma, V., Li, S.W., Galuba, W., Rabbat, M., Assran, M., Ballas, N.,
  Synnaeve, G., Misra, I., Jegou, H., Mairal, J., Labatut, P., Joulin, A.,
  Bojanowski, P.: Dinov2: Learning robust visual features without supervision.
  arXiv:2304.07193  (2023)

\bibitem{alpnet}
Ouyang, C., Biffi, C., Chen, C., Kart, T., Qiu, H., Rueckert, D.:
  Self-supervision with superpixels: Training few-shot medical image
  segmentation without annotation. In: Computer Vision--ECCV 2020: 16th
  European Conference, Glasgow, UK, August 23--28, 2020, Proceedings, Part XXIX
  16. pp. 762--780. Springer (2020)

\bibitem{alpnet_bp}
Ouyang, C., Biffi, C., Chen, C., Kart, T., Qiu, H., Rueckert, D.:
  Self-supervised learning for few-shot medical image segmentation. IEEE
  Transactions on Medical Imaging  \textbf{41}(7),  1837--1848 (2022).
  \doi{10.1109/TMI.2022.3150682}

\bibitem{Rani2023SSL}
Rani, V., Nabi, S.T., Kumar, M., Mittal, A., Kumar, K.: Self-supervised
  learning: A succinct review. Archives of Computational Methods in Engineering
   \textbf{30}(4),  2761--2775 (2023)

\bibitem{squeeze_and_excite}
Roy, A.G., Siddiqui, S., P{\"o}lsterl, S., Navab, N., Wachinger, C.: ‘squeeze
  \& excite’guided few-shot segmentation of volumetric images. Medical image
  analysis  \textbf{59},  101587 (2020)

\bibitem{shaharabany2023autosam}
Shaharabany, T., Dahan, A., Giryes, R., Wolf, L.: Autosam: Adapting sam to
  medical images by overloading the prompt encoder. arXiv preprint
  arXiv:2306.06370  (2023)

\bibitem{etis}
Silva, J., Histace, A., Romain, O., Dray, X., Granado, B.: Toward embedded
  detection of polyps in wce images for early diagnosis of colorectal cancer.
  International journal of computer assisted radiology and surgery  \textbf{9},
   283--293 (2014)

\bibitem{snell2017prototypical}
Snell, J., Swersky, K., Zemel, R.: Prototypical networks for few-shot learning.
  Advances in neural information processing systems  \textbf{30} (2017)

\bibitem{cvccolondb}
Tajbakhsh, N., Gurudu, S.R., Liang, J.: Automated polyp detection in
  colonoscopy videos using shape and context information. IEEE transactions on
  medical imaging  \textbf{35}(2),  630--644 (2015)

\bibitem{wang2020panet}
Wang, K., Liew, J.H., Zou, Y., Zhou, D., Feng, J.: Panet: Few-shot image
  semantic segmentation with prototype alignment. In: proceedings of the
  IEEE/CVF international conference on computer vision. pp. 9197--9206 (2019)

\bibitem{wu2023medical}
Wu, J., Ji, W., Liu, Y., Fu, H., Xu, M., Xu, Y., Jin, Y.: Medical sam adapter:
  Adapting segment anything model for medical image segmentation (2023)

\bibitem{Zhang2024SAM}
Zhang, Y., Shen, Z., Jiao, R.: Segment anything model for medical image
  segmentation: Current applications and future directions. Computers in
  Biology and Medicine  \textbf{171},  108238 (2024)

\end{thebibliography}

\end{document}